\def\year{2019}\relax
\documentclass[letterpaper]{article} 
\usepackage{aaai19}  
\usepackage{times}  
\usepackage{helvet}  
\usepackage{courier}  
\usepackage{url}  
\usepackage{graphicx}  
\usepackage{booktabs}
\usepackage{algorithm}
\usepackage{algorithmicx}
\usepackage{algpseudocode}

\usepackage{amsmath,bm}
\usepackage{amsfonts}
\usepackage{stfloats}
\usepackage{color}
\usepackage{indentfirst}

\nocopyright

\begin{document}

%
\title{Learning Segmentation Masks with the Independence Prior}
\author{Songmin Dai, Xiaoqiang Li\thanks{Corresponding author}, Lu Wang, Pin Wu, Weiqin Tong, Yimin Chen\\
School of Computer Engineering and Science, Shanghai University, China\\Shanghai Institute for Advanced Communication and Data Science, Shanghai University,  China\\\{laodar, xqli, luwang, wupin, wqtong, ymchen\}@shu.edu.cn}
\maketitle
\begin{abstract}
An instance with a bad mask might make a composite image that uses it look fake. This encourages us to learn segmentation by generating realistic composite images. To achieve this, we propose a novel framework that exploits a new proposed prior called the independence prior based on Generative Adversarial Networks (GANs). The generator produces an image with multiple category-specific instance providers, a layout module and a composition module. Firstly, each provider independently outputs a category-specific instance image with a soft mask. Then the provided instances' poses are corrected by the layout module. Lastly, the composition module combines these instances into a final image. Training with adversarial loss and penalty for mask area, each provider learns a mask that is as small as possible but enough to cover a complete category-specific instance. Weakly supervised semantic segmentation methods widely use grouping cues modeling the association between image parts, which are either artificially designed or learned with costly segmentation labels or only modeled on local pairs. Unlike them, our method automatically models the dependence between any parts and learns instance segmentation. We apply our framework in two cases: (1) Foreground segmentation on category-specific images with box-level annotation. (2) Unsupervised learning of instance appearances and masks with only one image of homogeneous object cluster (HOC). We get appealing results in both tasks, which shows the independence prior is  useful for instance segmentation and it is possible to unsupervisedly learn instance masks with only one image.
\end{abstract}

\section{Introduction}
Deep Convolutional Neural Networks (DCNNs) have been widely used and have achieved remarkable success in supervised semantic segmentation \cite{long2015fully,Chen2014Semantic,Ronneberger2015U,BU2LearnCRF}. But their training needs a large amount of images with pixel-level annotation. Such annotation is much more time consuming than box-level or image-level annotation, especially for the images of homogeneous object cluster (HOC) \cite{HOC} because of densely distributed objects with various degrees of occlusion. Is it possible to learn image segmentation with low-cost annotation or even just one image of HOC without any annotation? The answer suggested by our work is yes.

A natural image consists of multiple separated coherent objects. And objects often appear, move and change independently. It seems that the image segmentation task is just to group the coherent pixels into independent components, which can be treated as a independent component analysis (ICA) or blind source separation problem. But thoughts like grammar of images \cite{zhu2007stochastic}, scene graph \cite{newell2017pixels,johnson2018image} and geometric relationship \cite{hu2018relation,lin2018st} imply that objects seeming disentangled are actually weakly interrelated by their poses (position, scale and orientation) at larger spatial scales. To formalize and better understand these intuitions, a graphical model for scene images is developed firstly. Then we point out that the real independent factors among objects are their appearances. Actually the appearances of objects are conditionally independent given the categories of objects. We call it as the independence prior.
\begin{figure}[]
\centering
\includegraphics[width=\linewidth,trim={0cm 0cm 0cm 0cm},clip]{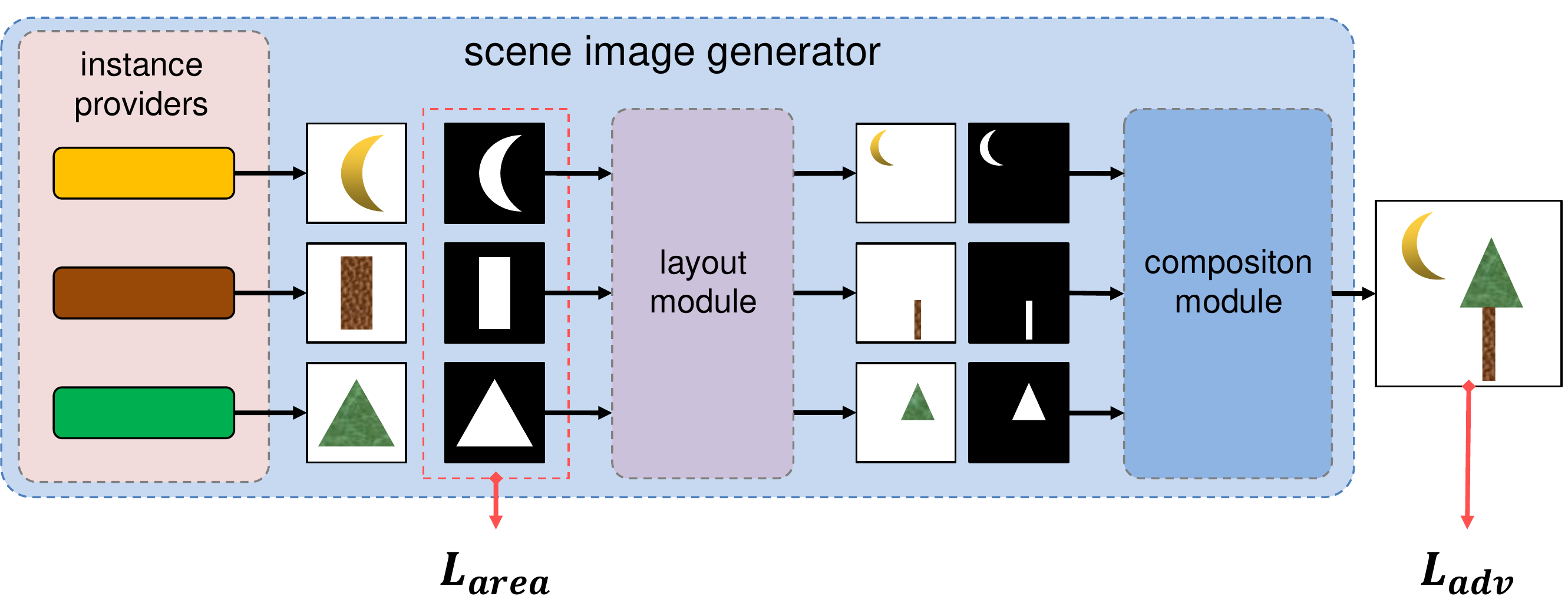}
\caption{Our GAN based framework generates scene images by compositing instances with independent appearances and specific categories, which is trained with adversarial loss and mask area penalty.}
\label{outline}
\end{figure}

We argue that the independence prior is important for building a strong connection between the quality of a composite image and the quality of the instance masks it uses. To exploit it for unsupervised instance segmentation, we propose a framework based on generative adversarial networks (GANs) \cite{Goodfellow2014Generative}, which learns instance segmentation by optimizing the composite images from instances with specific categories, independent appearances, the smallest possible masks and suitable layout. Figure \ref{outline} shows an overview of the proposed framework. We apply our framework on two practical visual tasks. In the category-specific foreground segmentation task, there are two different category-specific instance providers. One is a trainable foreground instance provider adapted from a segmentation network, and the other is an untrainable random sampler on prepared backgrounds collected from natural images. Fake images are generated by pasting carved foregrounds from input images into various backgrounds. The foreground instance provider learns instance masks by struggling to find minimal necessary regions that keep foreground instances unbroken and realistic. We use data distillation \cite{distillation} for the raw result to improve performance, which is useful when the training data is very limited. In the task of unsupervised learning of appearances and masks with only one image of HOC, there are many enough trainable instance providers with shared parameters to generate instances for dense homogeneous objects. Fake images are generated by compositing them with structure-free layouts. With reducing the artifacts and mask area penalty, learned appearances gradually get improved and each mask finally just covers a region that is as small as possible but enough for exactly one object.

Our main contributions are as follows:
\begin{itemize}
\item We formalize intuitions about natural images with a graphical model, which captures the dependence of appearance on category, the conditional independence of appearances and the geometric co-occurrence of objects.
\item We propose a novel framework that learns instance appearances and segmentation masks with the independence prior.
\item Our first application provides a new feasible solution for foreground segmentation on category-specific images with box-level annotation.
\item Our second application shows the possibility of learning instance appearances and segmentation masks from just one image of HOC without any annotations. 
\end{itemize}

\section{Related Work}
\subsubsection{Localization cues based on the dependence between image part and instance category.}Methods like \cite{Zhou2016Learning,selvaraju2017grad} use learned classification networks to find out the image parts with category-specific patterns. But such localization cues only help us to find sparse and local regions with discriminative features, which rely on the dependence between image part and instance category. \cite{Wei2017Object} tried to gradually erase as much as possible discriminative region contributing for classification through a progressive trained classifier. But region without discriminative features may still be missing without their postprocessing.
\subsubsection{Grouping cues based on the dependence between image parts.}Intra-object parts have a high affinity with each other, but it is much lower for the affinity between inter-object parts. We think such prior about the dependence between image parts is important for bottom up grouping and instance segmentation.  Recent weakly supervised semantic segmentation methods widely use grouping cues that are built in CRF, MCG, GrabCut, ect. techniques. \cite{BU1,BU2LearnCRF,BU3GrabCut,BU4MCG}. But untrainable CRF, MCG and GrabCut are human designed grouping prior based on color, texture etc. appearance, which is unlearnable. Trainable CRF is possible to learn the dependence of image parts without human designed potential functions, but costly segmentation labels are necessary \cite{BU2LearnCRF}. \cite{ahn2018learning} tried to learn affinities between image parts with image-level supervision. But they modeled such affinities only on local pairs. \cite{Pathak2017Learning} models the dependence of image parts through the motion association of image parts from videos. By using motion cue to correct the grouping based on low-level appearance, their experiments show that this indeed improves the segmentation performance.
\subsubsection{Layered generative models for images.}Some generative models use image layers to composite a scene image that contains multiple objects, which can get some benefits from modeling each object separately. In the works of \cite{johnson2018image,Yang2017LR,Vondrick2016Generating}, generators can be trained to produce a coarse soft mask for each instance without pixel-level annotation. However, these generators didn't exploit the independence prior, image layers may communicate with and thus compensate each other to reduce the artifacts of composite results, which hinders the optimization of soft masks. In \cite{Eslami2016Attend,huang2015efficient}, it was assumed that both the appearances and poses of objects are factorized, VAE  \cite{Kingma2013Auto} was used to inference each layer and reconstruct the image by compositing the inferenced layer. In Tagger \cite{Greff2016Tagger}, a RNN model was designed to progressively group coherent pixels and separate the independent objects, which was supervised via a denoise task. It was assumed that removing the interference between different objects can lead to a better denoise performance. But what are the real independent factors was still not pointed out, and the dependence of these factors are not minimized directly.
\subsubsection{Minimizing the dependence between factors with generative models.} Our framework forces generator to take the independence prior by compositing instances with independent appearances. In the foreground segmentation tasks, we paste the carved foreground into a different background sampled independently, which is inspired by the resampling operation proposed in \cite{brakel2017learning}. They minimized mutual information by using GAN to minimize the f-divergence \cite{Nowozin2016f} between joint distribution and product of its marginals. A more formally expression can be found in Mutual Information Neural Estimator (MINE) \cite{Belghazi2018MINE}. 

\section{The Independence Prior}
\label{headings}
\begin{figure}[!htbp]
  \centering
  \includegraphics[width=0.12\textwidth]{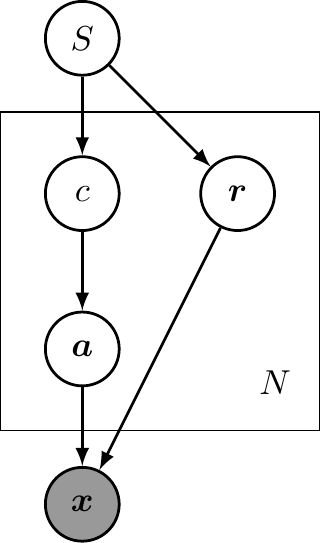}
  \caption{A graphical model for scene images that contain $N$ instances.}
  \label{PGM_org}
\end{figure}

In this paper, we hierarchically decompose an image into three level representations: part (pixels, feature, patch, etc. visual primitives), instance (single object) and image. And we use a corresponding description for an image with hierarchical latent factors: \textit{appearance} (both appearance and surface mask) for the pixel level, object category and pose for the instance level and scene for the image level.
Without loss of generality, we assume that a natural scene image is the result of compositing $N$ instances according to the scene-specific layout and their visible order. Images with $M<N$ instances can be treated as a special case with $N-M$ invisible instances. More formally, a natural image $\bm{x}$ of scene $S$ is generated from $N$ instances with categories $c_1,c_2,...,c_N$, category-specific \textit{appearances} $\bm{a}_1,\bm{a}_2,...,\bm{a}_N$ and scene-specific poses $\bm{r}_1,\bm{r}_2,...,\bm{r}_N$ according to their stacking orders $1,2,...,N$:
\begin{equation}
p(\bm{x})=\int_{\bm{z}} p(\bm{x}|\bm{z})p(\bm{z})\,d\bm{z}\,,
\end{equation}
where $\bm{z}$ denotes all of the above latent factors,
\begin{equation} p(\bm{z})=p(c_1,\bm{a}_1,\bm{r}_1,c_2,\bm{a}_2,\bm{r}_2,...,c_N,\bm{a}_N,\bm{r}_N,S)\,.
\end{equation}
Based on our hierarchical decomposition, $p(\bm{z})$ can be further factorized into:
\begin{equation}
p(S)p(c_1,\bm{r}_1,c_2,\bm{r}_2,...,c_N,\bm{r}_N|S)\prod_{i=1}^N p(\bm{a}_i|c_i)\,,
\end{equation}
where $p(c_1,\bm{r}_1,c_2,\bm{r}_2,...,c_N,\bm{r}_N|S)$ term models the scene-specific geometric co-occurrence association of instances and captures the dependence of object categories and their poses. The \textit{appearance} of each instance depends only on its category but not any other. The factorization term $\prod_{i=1}^N p(\bm{a}_i|c_i) $ shows the conditional independence of instance \textit{appearances}. See
Figure \ref{PGM_org} for a graphical model depiction. 

We can summarize the following three priors about the dependence between these factors:
\begin{itemize}
\item \textbf{The dependence between image part and instance category} The appearance of an object part is dependent on its object category. 
\item \textbf{The dependence between image parts} The appearances of inter-object parts are independent of each other given object categories, but intra-object parts are still related to each other. Intuitively, there is a much higher affinity between intra-object parts than inter-object parts.
\item \textbf{The dependence between instance poses}  Objects of a specific scene are related to each other by the geometric co-occurrence association, such as the coexistence of eye glasses and human face, fish and water, car and road, etc.
\end{itemize}

These properties lead to following factors that mainly affect the realistic of a composite image:
\begin{itemize}
\item \textbf{Appearance} The appearance of image parts should obey the statistical association corresponding to its category so that we can't find any artifacts from it.
\item \textbf{Segmentation mask} 
The mask of an instance for image composition shouldn't break down the integrity of its surface so that the dependence between image parts won't be violated when the carved surface is combined with new surfaces with independent appearances.
\item\textbf{Geometric relationship} 
Instances should be placed with suitable poses according to the geometric co-occurrence association between instances.
\end{itemize}
\begin{figure}[!h]
  \centering
  \begin{minipage}[c]{0.15\textwidth}
  \centering
  \includegraphics[width=\textwidth]{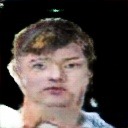}\\
  \end{minipage}
  \begin{minipage}[c]{0.15\textwidth}
  \centering
  \includegraphics[width=\textwidth]{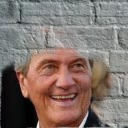}\\
  \end{minipage}
  \begin{minipage}[c]{0.15\textwidth}
  \centering
  \includegraphics[width=\textwidth]{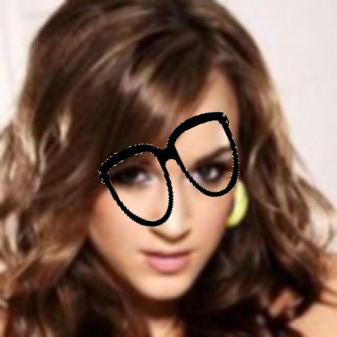}\\
  \end{minipage}
  \caption{Images with artifacts caused by different factors. Left: Artifacts comes from the appearance. Middle: Artifacts comes from segmentation mask of the face. Right: Artifacts comes from the unsuitable placement of the eye glasses on that face.}
  \label{fg1}
\end{figure}

They imply that it is possible to train generative models to jointly learn object appearances, segmentation masks and geometric relationships unsupervisedly. But jointly learning to place multiple objects and inference segmentation masks is still hard for high resolution images as far as we know, actually ST-GAN \cite{lin2018st} shows that problems may exist even if only two objects with perfect appearances and masks need to be arranged. Our framework only focuses on the special case where layout $\bm{r}_1,\bm{r}_2,...,\bm{r}_N$ is easy to design by hand and object categories $c_1,c_2,...,c_N$ are known and fixed. In such cases, artifacts caused by unsuitable layout can be avoided, and object appearances are independent of each other because the $\bm{a}_1,\bm{a}_2,...,\bm{a}_N$ are d-separated by $c_1,c_2,...,c_N$. With this independence prior, we propose a framework that learns object appearances and segmentation masks by reducing artifacts like the first two examples showed in Figure \ref{fg1}.

\begin{figure*}[h]
\centering
\includegraphics[width=\textwidth]{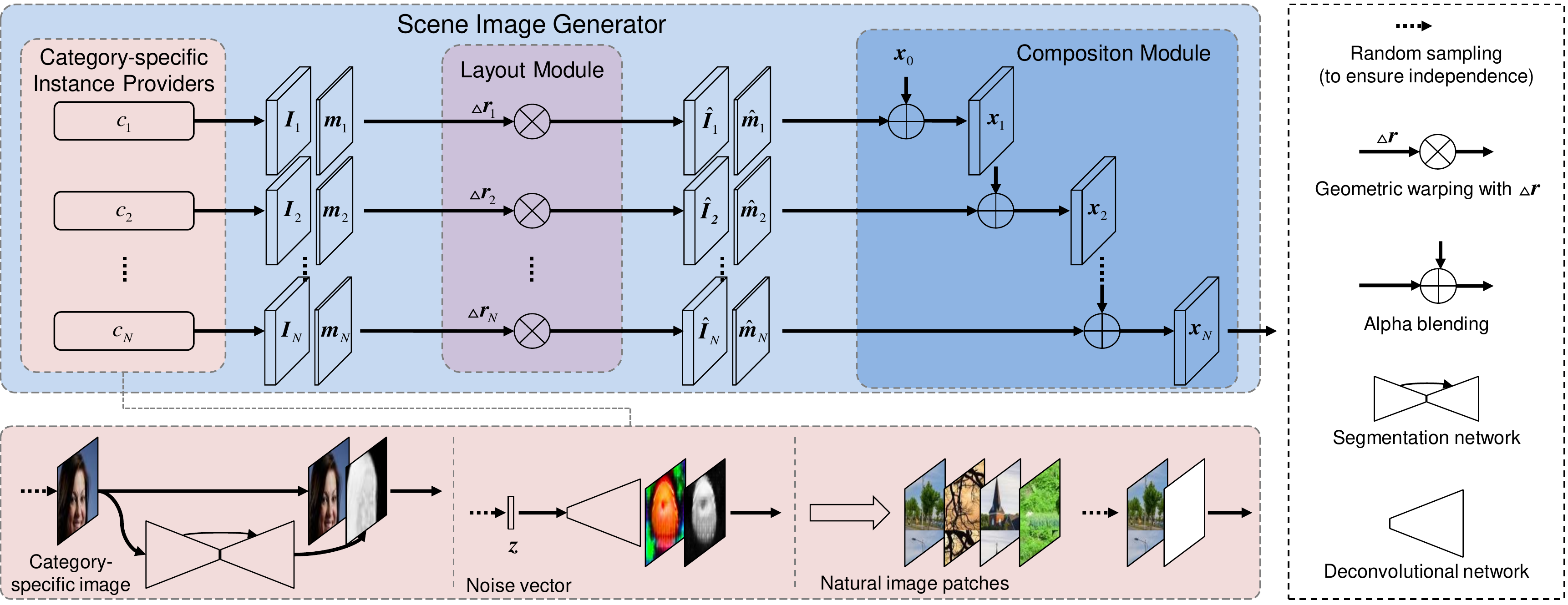}
\caption{Generator architecture in the proposed framework. The bottom shows details of three implementation cases of instance providers which are used in our applications. The first case produces instances from randomly sampled category-specific image with a segmentation network, the second case produces instances from randomly sampled noise vector with a deconvolutional network and the last case produces instances from randomly cropped natural image patches.}
\label{overview}
\end{figure*}

\section{Our Framework}
To learn object appearances and segmentation masks by generating realistic composite images of specific scene with GANs. We need to avoid the image layers communicating with and compensating each other so that the artifacts of each instance can be effectively showed in the final composite image. And the discriminator shouldn't win easily by finding out the artifacts caused by layout, which discourages the discriminator from finding more artifacts caused by instance masks. As showed in Figure \ref{overview}, our model generates a scene-specific image by compositing multiple instances of categories $c_1,c_2,...,c_N$ with a suitable layout. To ensure that object appearances are independent of each other, the instance images for compositing are sampled independently from category-specific instance providers. To place objects in a suitable geometric relationship, geometric warpings are used for instance images to correct their poses appearing in the final composite images according to scene-specific layout priors. Such priors could be summarized based on human observations or adapted from bounding box annotations of scene images like \cite{johnson2018image}.
\subsection{Generator for Scene Images}
\subsubsection{Category-specific Instance Providers}
Category-specific instance providers are designed to independently output images of four channels for categories $c_1,c_2,...,c_N$ according to specific tasks, the first three channels RGB capture the color appearance, the last channel Alpha represents the soft mask. Any architecture that produces category-specific instances can be implemented according to our requirements. Figure \ref{overview} shows some implementation cases of instance providers, which are used and described in our applications. More details can be found in the application section.
\subsubsection{Layout Module}
Because the instances are outputted independently, they are also forced to own independent poses $\bm{r}_1,\bm{r}_2,...,\bm{r}_N$. So geometric warpings are needed to correct them into a suitable layout. The geometric warping parameters $\Delta \bm{r}_1, \Delta \bm{r}_2, ..., \Delta \bm{r}_N$ are shared among RGBA channels:
\begin{equation}
\begin{aligned}
\hat{\bm{I}}_i=&ST(\bm{I}_i,\Delta \bm{r}_i)\\
\hat{\bm{m}}_i=&ST(\bm{m}_i,\Delta \bm{r}_i)\,,\\
\end{aligned}
\end{equation}
where $\hat{\bm{I}}_i$ is the warping result of $i$th layer instance's appearance  $\bm{I}_i$(the RGB channels), $\hat{\bm{m}}_i$ is the warping result of the $i$th layer instance's soft mask $\bm{m}_i$ (the Alpha channel), $ST$ is the geometric warping operator.
\subsubsection{Composition Module}
We composite a complex image that contains many instances by recursively applying Alpha blending according to their stacking orders:
\begin{equation}
\bm{x}_i = \hat{\bm{I}}_i\odot \hat{\bm{m}}_i + \bm{x}_{i-1}\odot (1-\hat{\bm{m}}_i)\,,
\end{equation}
where the $\bm{x}_{i}$ is the composite image of the first $i$ layer instances, $\bm{x}_0$ is set to be a pure black image.
\subsection{Adversarial Training with Mask Area Penalty}
 The key for instance providers is to provide instance images independently and thus break down the statistical association among them. This encourages instance providers not to separate high affinity parts and output complete surfaces so that artifacts will not be shown when composite with irrelevant neighbor pixels. 
We optimize the generator with adversarial training, but there are trivial solutions if only a adversarial loss is used. The instance outputted by a provider maybe contain completed surfaces of multiple objects even a complete scene image in order to avoid breaking down any surfaces and showing any artifacts. So a mask area penalty is used for encouraging instance providers to output a minimal necessary surface region that covers a single complete instance. Intuitively, instance providers struggle to remove parts with low affinities to necessary discriminative parts and group together the ones with high affinities. We train the discriminator $D$ and the generator $G$ by following loss functions:
\begin{equation}
L_D=\mathbb{E}_{\bm{x}}[\log(D(\bm{x}))]+\mathbb{E}_{\bm{x}_g}[\log(1-D(\bm{x}_g))]
\end{equation}
\begin{equation}
L_G = \mathbb{E}_{\bm{x}_g}[\log(D(\bm{x}_g))]+\lambda L_{area}\,,
\end{equation}
where $L_{area}=\mathbb{E}_{\bm{m}}[\frac{1}{N}\sum_i^N(\frac{\|\bm{m}_i\|_1}{A}-a)^2]$, $\bm{x}$ and $\bm{x}_g$ are drawn from real data and fake composite distributions, $\bm{m}_i$ is the soft mask outputted from the $i$th layer instance provider, $A$ is the image area in pixels, $a$ is an empirical constant for the smallest mask area. 
\section{Two Applications}\label{applications}
\subsection{Foreground Segmentation on Category-specific Images}
Firstly we apply our framework on category-specific foreground segmentation tasks with box-level annotation. In this task, all images are preprocessed by cropping with box annotations. Every image contains at least one foreground of the target category. Actually there are only two instances with known and fixed categories and stacking orders. We use two instance providers to independently provide the foreground instances and background instances.\subsubsection{Foreground Provider}
The foreground provider is implemented with a segmentation network. As shown in the first implementation case of Fig. \ref{overview}, a foreground instance is constructed by concatenating the input RGB image and the corresponding inferred foreground mask, where we assume that foreground objects are almost always showed with little occlusion.
\subsubsection{Background Provider}Generating or inferencing high-resolution backgrounds (by image inpainting) may be hard due to the variety of appearance and the non-negligible occlusion. But it is much easier to select good enough image as background, like crawl context related images from web or sample enough patches without any foreground region from the original given images. So we construct background instances by sampling patches from scene-specific natural images, and their masks just cover all the pixels of such patches, as shown in the third implementation case of Figure \ref{overview}.
\subsubsection{Layout Prior}
It is assumed that the dependence between foregrounds and backgrounds can be reduced to weak enough after cropping by bounding boxes, which means the instance poses are also independent of each other. So no geometric corrections are needed, and $\Delta \bm{r}_i=\bm{0}$.

\subsection{Learning Instance Appearances and Masks From One Image of HOC}
\begin{figure}[!htb]
\centering
\includegraphics[width=0.22\textwidth]{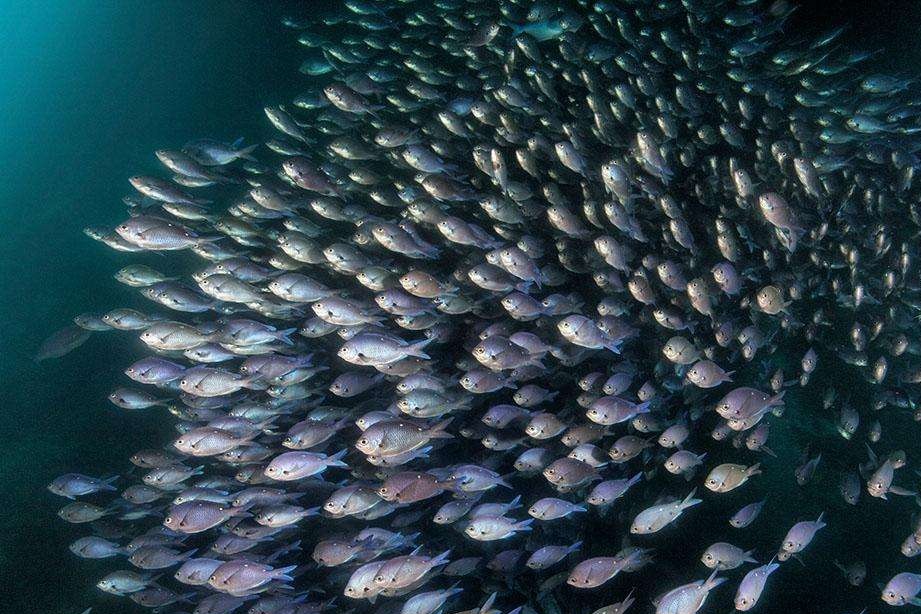} \includegraphics[width=0.22\textwidth]{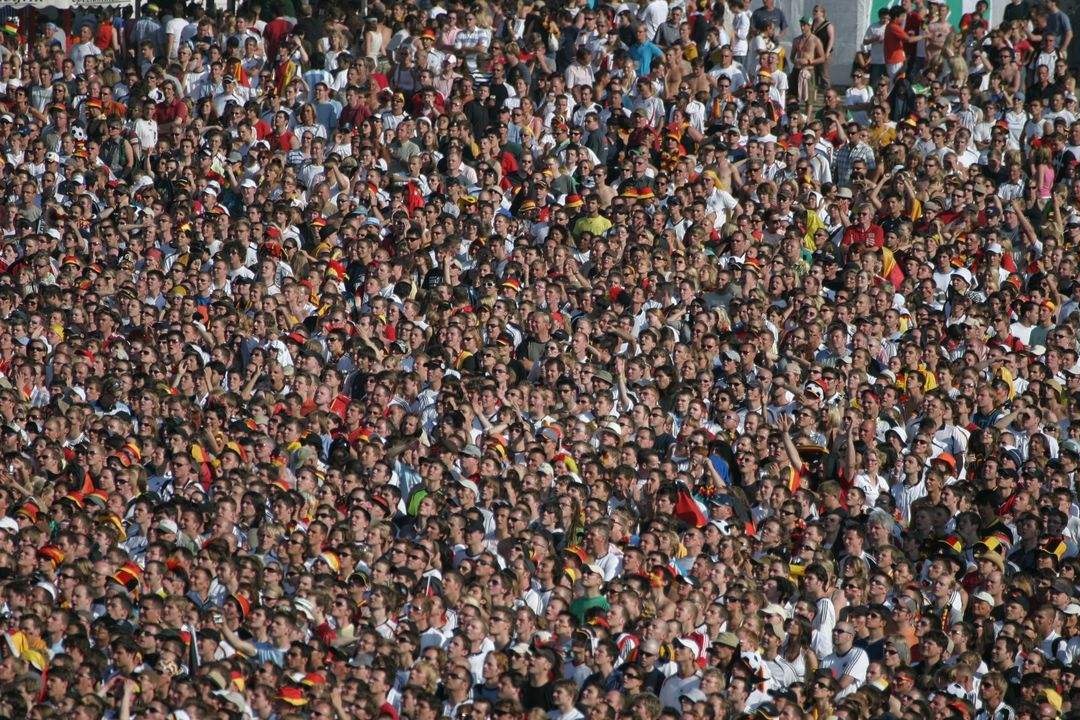}\\\includegraphics[width=0.22\textwidth]{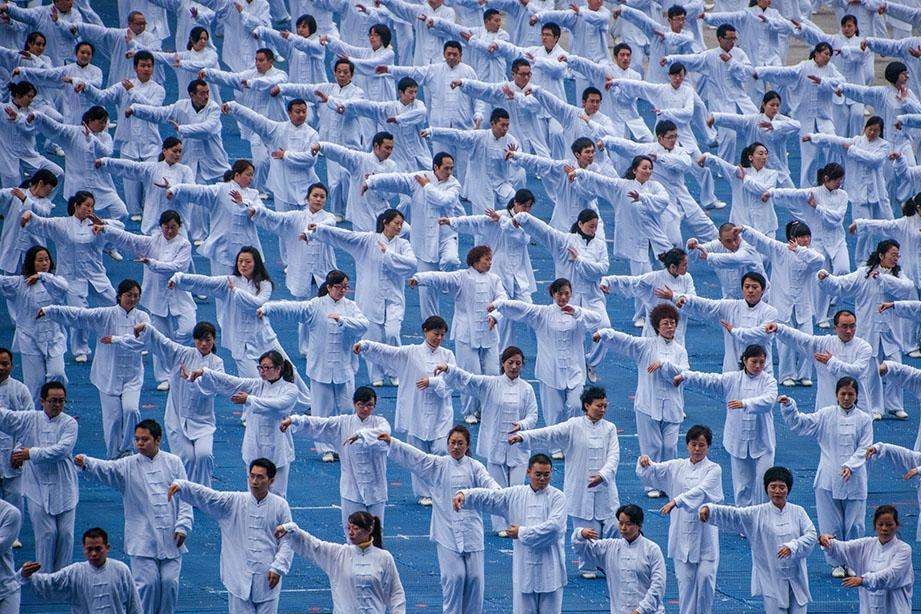} \includegraphics[width=0.22\textwidth]{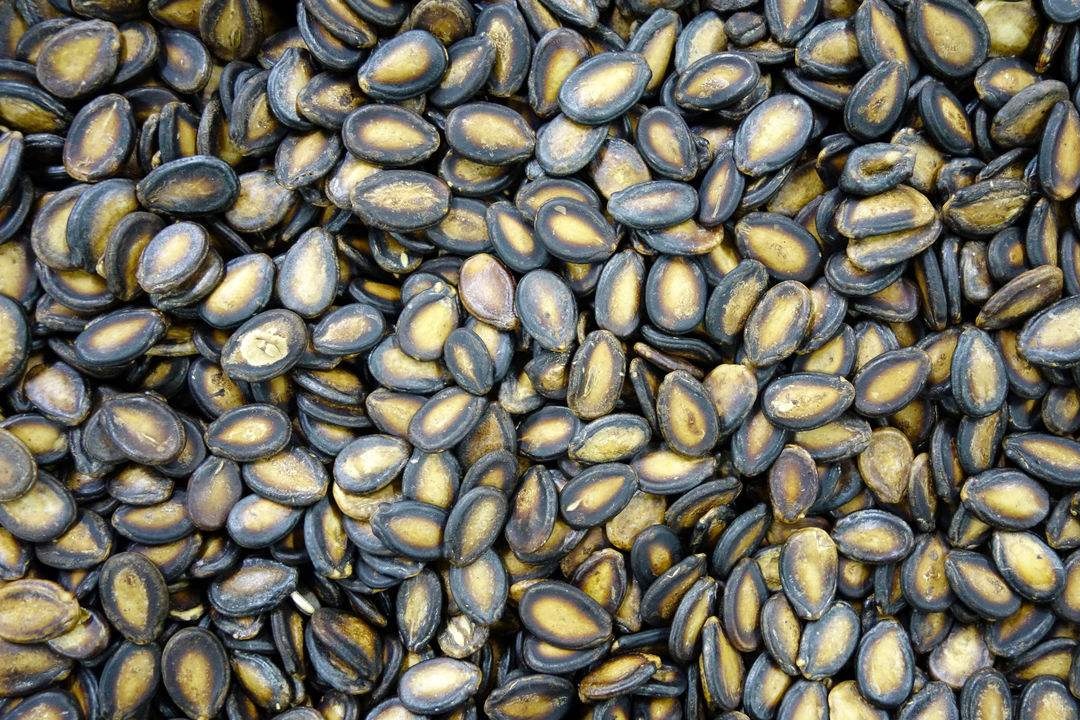}\\
\caption{Some images of HOC. The right column is the special case, images of DSF-HOC.}
\label{fg_crowded_images}
\end{figure}
Sometimes we have images of HOC that contains a large number of foreground objects with the same category and diversity appearances. Figure\ref{fg_crowded_images} shows some of them. 
Theoretically, one image of HOC contains a sufficient amount of information about the appearance of foreground objects. So it is possible to learn the appearances, segmentation masks of foreground objects from just one image of HOC even if many of them are overlapped. To efficiently demonstrate this possibility, we apply our framework on the special cases of HOCs, images of dense structure-free homogeneous object cluster (DSF-HOC), because of the simplicity of instance layout and the invisibility of background, as showed in the right column of Figure \ref{fg_crowded_images}. 

We generate small image patches but not a whole image, main reasons are: 1. We can collect much more training samples rather than only a single one image; 2. We can greatly reduce the size of image layers $N$; 3. Smaller input size is required for maintaining the visual quality of instances.

\subsubsection{Foreground Providers}
To learn the appearance and object mask of foreground objects, we provide foreground instances by a trainable generator that maps a random noise vector $\bm{z}$ to a RGBA image as shown in the second implementation case of Figure \ref{overview}. All of $N$ foreground providers share the same generator because of the category uniformity, but sample $\bm{z}$ independently to guarantee the independence between appearances. 
\subsubsection{Background Provider} We construct background images for DSF-HOC patches by just sampling from pure black images or real patches with missing regions. The foreground providers are forced to generate complete foreground objects to cover them. 
\subsubsection{Layout Prior}
Because of the structure-free property of DSF-HOCs, instances appear randomly and independently with allowed poses that range from $\bm{r}_{min}$ to $\bm{r}_{max}$. It is expected that the learned instances own canonical poses, so $\Delta \bm{r}_i$ is sampled from uniform distribution $U(\bm{r}_{min},\bm{r}_{max})$. 

\newcommand{\myFigure}[2]{\includegraphics[width=#1\linewidth]{#2}}

\begin{figure*}[htbp]
\centering
\linespread{0.0}\selectfont
\begin{tabular}{c@{\hspace{0.002\linewidth}}c@{\hspace{0.002\linewidth}}c@{\hspace{0.002\linewidth}}c@{\hspace{0.002\linewidth}}c@{\hspace{0.002\linewidth}}c@{\hspace{0.002\linewidth}}c@{\hspace{0.002\linewidth}}c@{\hspace{0.002\linewidth}}c@{\hspace{0.002\linewidth}}c@{\hspace{0.002\linewidth}}c@{\hspace{0.002\linewidth}}
c@{\hspace{0.002\linewidth}}c@{\hspace{0.002\linewidth}}
c@{\hspace{0.002\linewidth}}c}
\myFigure{0.06}{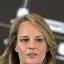}&\myFigure{0.06}{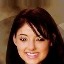}&\myFigure{0.06}{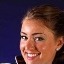}&\myFigure{0.06}{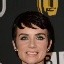}&\myFigure{0.06}{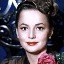}&\myFigure{0.06}{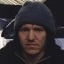}&\myFigure{0.06}{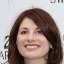}&\myFigure{0.06}{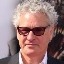}&\myFigure{0.06}{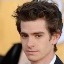}&\myFigure{0.06}{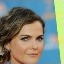}&\myFigure{0.06}{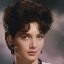}&\myFigure{0.06}{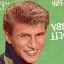}&\myFigure{0.06}{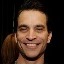}&\myFigure{0.06}{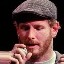}&\myFigure{0.06}{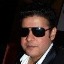}\\
\myFigure{0.06}{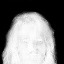}&\myFigure{0.06}{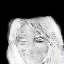}&\myFigure{0.06}{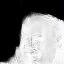}&\myFigure{0.06}{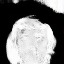}&\myFigure{0.06}{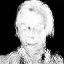}&\myFigure{0.06}{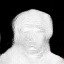}&\myFigure{0.06}{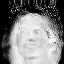}&\myFigure{0.06}{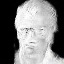}&\myFigure{0.06}{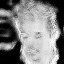}&\myFigure{0.06}{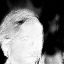}&\myFigure{0.06}{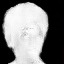}&\myFigure{0.06}{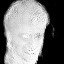}&\myFigure{0.06}{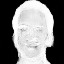}&\myFigure{0.06}{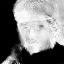}&\myFigure{0.06}{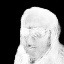}\\
\myFigure{0.06}{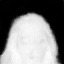}&\myFigure{0.06}{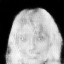}&\myFigure{0.06}{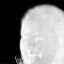}&\myFigure{0.06}{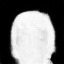}&\myFigure{0.06}{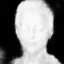}&\myFigure{0.06}{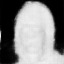}&\myFigure{0.06}{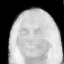}&\myFigure{0.06}{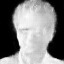}&\myFigure{0.06}{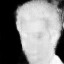}&\myFigure{0.06}{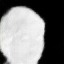}&\myFigure{0.06}{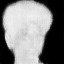}&\myFigure{0.06}{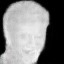}&\myFigure{0.06}{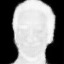}&\myFigure{0.06}{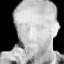}&\myFigure{0.06}{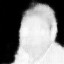}\\
\end{tabular}
\linespread{0.0}\selectfont
\begin{tabular}{c@{\hspace{0.002\linewidth}}c@{\hspace{0.002\linewidth}}c@{\hspace{0.002\linewidth}}c@{\hspace{0.002\linewidth}}c@{\hspace{0.002\linewidth}}c@{\hspace{0.002\linewidth}}c@{\hspace{0.002\linewidth}}c@{\hspace{0.002\linewidth}}c@{\hspace{0.002\linewidth}}c@{\hspace{0.002\linewidth}}c@{\hspace{0.002\linewidth}}
c@{\hspace{0.002\linewidth}}c@{\hspace{0.002\linewidth}}
c@{\hspace{0.002\linewidth}}c}
\myFigure{0.06}{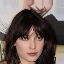}&\myFigure{0.06}{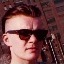}&\myFigure{0.06}{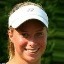}&\myFigure{0.06}{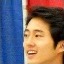}&\myFigure{0.06}{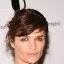}&\myFigure{0.06}{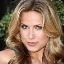}&\myFigure{0.06}{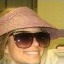}&\myFigure{0.06}{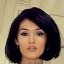}&\myFigure{0.06}{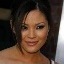}&\myFigure{0.06}{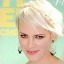}&\myFigure{0.06}{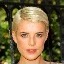}&\myFigure{0.06}{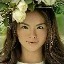}&\myFigure{0.06}{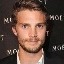}&\myFigure{0.06}{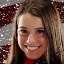}&\myFigure{0.06}{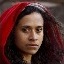}\\
\myFigure{0.06}{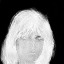}&\myFigure{0.06}{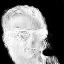}&\myFigure{0.06}{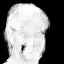}&\myFigure{0.06}{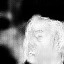}&\myFigure{0.06}{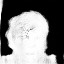}&\myFigure{0.06}{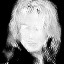}&\myFigure{0.06}{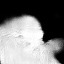}&\myFigure{0.06}{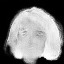}&\myFigure{0.06}{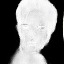}&\myFigure{0.06}{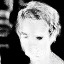}&\myFigure{0.06}{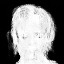}&\myFigure{0.06}{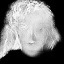}&\myFigure{0.06}{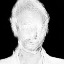}&\myFigure{0.06}{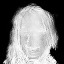}&\myFigure{0.06}{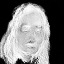}\\
\myFigure{0.06}{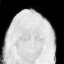}&\myFigure{0.06}{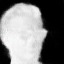}&\myFigure{0.06}{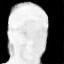}&\myFigure{0.06}{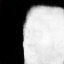}&\myFigure{0.06}{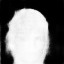}&\myFigure{0.06}{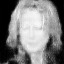}&\myFigure{0.06}{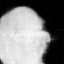}&\myFigure{0.06}{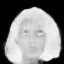}&\myFigure{0.06}{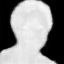}&\myFigure{0.06}{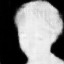}&\myFigure{0.06}{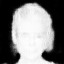}&\myFigure{0.06}{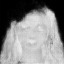}&\myFigure{0.06}{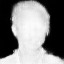}&\myFigure{0.06}{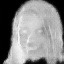}&\myFigure{0.06}{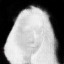}\\
\end{tabular}
  \caption{Foreground segmentation results on CelebA dataset. Soft masks are showed. The first and fourth rows are real images, the second and fifth rows are the corresponding images of the raw results, the third and six rows are the corresponding images of the results with data distillation. Ground truth is unavailable.}
  \label{face_result}
\end{figure*}

\section{Experiments and Results}\label{exp}
\subsection{Experimental Settings}
In all experiments, we use TensorFlow \cite{Abadi2016TensorFlow} library to build and train our models. Discriminators in our experiments are designed based on the SN-GAN\cite{Miyato2018Spectral}. 
All the other neural networks use batch normalization in most of layers. The segmentation network is adopted from UNet \cite{Ronneberger2015U}. Generators in the DSF-HOC tasks are designed based on deep deconvolution neural networks. All models are trained with the ADAM \cite{Kingma2014Adam} optmizer with $\beta_1=0.5$ and $\beta_2=0.999$. The learning rate is fixed as 0.0002 for discriminators and 0.0004 for generators. See Appendix for further details. 
\subsection{Foreground Segmentation on Category-specific Images}
\begin{figure}[htb]
  \centering
  \includegraphics[width=\linewidth]{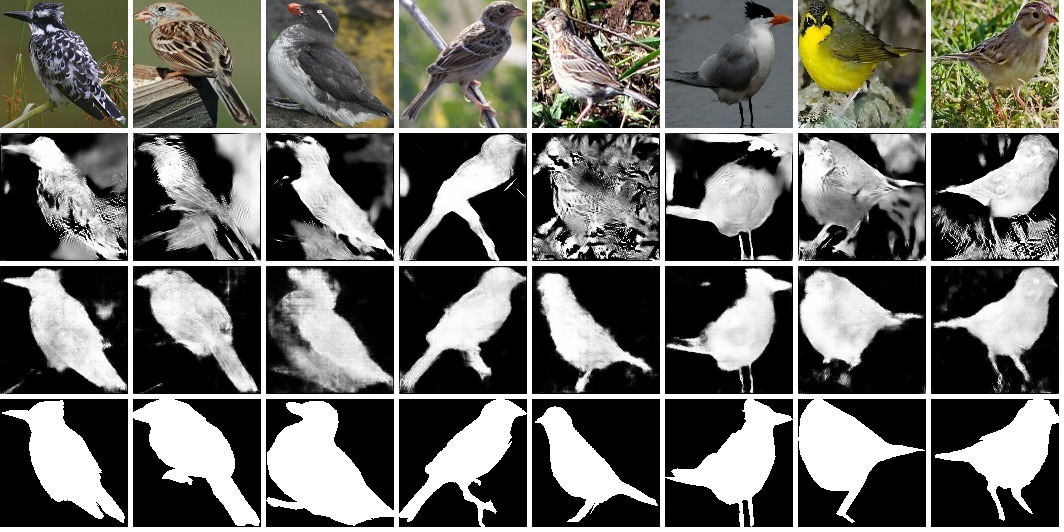}
  \caption{Foreground segmentation results on Caltech-200 bird dataset. Soft masks are showed. From top to bottom, the image rows are real images, raw results, results with data distillation and the ground truth.}
 \label{bird_result}
\end{figure}
We evaluate our foreground segmentation method on the CelebA \cite{liu2015faceattributes} and Caltech-200 bird \cite{wah2011caltech} datasets. The CelebA dataset contains 202,599 face images with box annotation. Caltech-200 bird dataset contains 11,788 images with both box-level and pixel-level annotation. 
\subsubsection{Data Preparation.}
Firstly we crop images using provided bounding boxes. Then cropped images are scaled to size $128\times 128$ as category-specific images. We construct the reference background images for CelebA by sampling patches with the same size from INRIAPerson \cite{Dalal2005Histograms} dataset's negative images but from images crawled from the web with keywords 'landscape' or 'tree' for Caltech-200 bird because of more tree and landscape background appearing in bird images.
\subsubsection{Raw Results.} The results by adversarial training are shown in the first rows under input images of Figure \ref{face_result} and Figure \ref{bird_result}. The segmentation model with adversarial training tries to infer a minimal region that covers a integral foreground object. There is no try to search architectures for segmentation networks and discriminators, but satisfactory segmentation masks are obtained on CelebA dataset. There are weird noises especially for the Caltech-200 bird dataset. These noises seem to be nothing to do with the texture or with explainable shapes. This may be caused by: 1. Intrinsic problems of adversarial training; 2. Training samples are not enough for showing the foreground and background independent effectively, enough cases with similar foregrounds appear in various backgrounds are needed; 3. Training samples may be not enough to show the generative appearance. We believe that intrinsic problems of adversarial training can be avoided with better adversarial training methods in the future. Problems caused by dataset size could be solved by finding more bird images from the web and cropping with a bird detector.
\subsubsection{Results with Data Distillation.}
We show that the raw result can be easily improved by data distillation in the second rows under input images Figure \ref{face_result} and Figure \ref{bird_result}. With data distillation, we get appealing results on both dataset. We compute the mIOU score on Caltech-200 bird dataset using the ground-truth, and the final result gets performance of 0.78. Qualitative evaluation on CelebA dataset is hard because pixel-level annotation is missed and costly. But a visually accuracy for semantic boundary can be observed clearly. The segmentation results on Caltech-200 bird dataset show the powerful unsupervised grouping ability, even though there is highly similar texture between the background and foreground, our method can segment out a bird shape mask which is very close to the ground truth. We argue that it is hard for discriminative methods because of lacking ability to model generative detail, part-part dependency and shape structure. And the soft mask generated by our method will be more suitable for image synthesis because of the direct optimization of the composite quality.
\begin{figure*}[!htbp]
\centering
\linespread{0.3}\selectfont
\resizebox{\textwidth}{!}{
\begin{tabular}{c@{\hspace{0.005\linewidth}}c@{\hspace{0.005\linewidth}}c}
\includegraphics[width=0.33\linewidth]{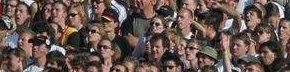}&\includegraphics[width=0.33\linewidth]{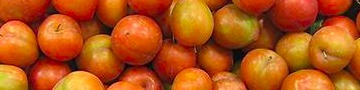}& \includegraphics[width=0.33\linewidth]{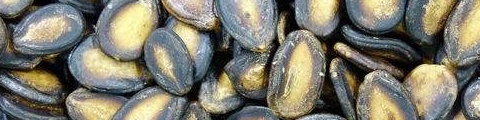}\\{\begin{tabular}{c@{\hspace{0.005\linewidth}}c@{\hspace{0.005\linewidth}}c@{\hspace{0.005\linewidth}}c@{\hspace{0.005\linewidth}}c}
\myFigure{0.062}{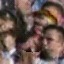}&\myFigure{0.062}{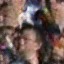}&\myFigure{0.062}{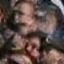}&\myFigure{0.062}{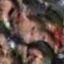}&\myFigure{0.062}{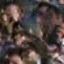}\\\myFigure{0.062}{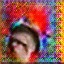}&\myFigure{0.062}{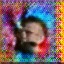}&\myFigure{0.062}{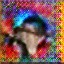}&\myFigure{0.062}{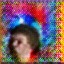}&
\myFigure{0.062}{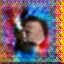}\\\myFigure{0.062}{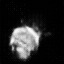}&\myFigure{0.062}{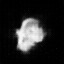}&\myFigure{0.062}{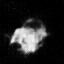}&\myFigure{0.062}{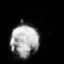}&
\myFigure{0.062}{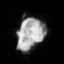}\\\end{tabular}}&{\begin{tabular}{c@{\hspace{0.005\linewidth}}c@{\hspace{0.005\linewidth}}c@{\hspace{0.005\linewidth}}c@{\hspace{0.005\linewidth}}c}\myFigure{0.062}{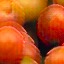}&\myFigure{0.062}{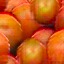}&\myFigure{0.062}{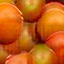}&\myFigure{0.062}{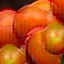}&\myFigure{0.062}{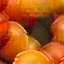}\\
\myFigure{0.062}{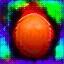}&\myFigure{0.062}{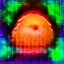}&\myFigure{0.062}{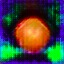}&\myFigure{0.062}{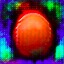}&
\myFigure{0.062}{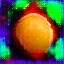}\\\myFigure{0.062}{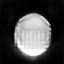}&\myFigure{0.062}{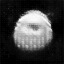}&\myFigure{0.062}{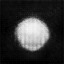}&\myFigure{0.062}{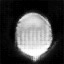}&
\myFigure{0.062}{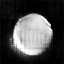}\\\end{tabular}}&{\begin{tabular}{c@{\hspace{0.005\linewidth}}c@{\hspace{0.005\linewidth}}c@{\hspace{0.005\linewidth}}c@{\hspace{0.005\linewidth}}c}\myFigure{0.062}{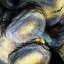}&\myFigure{0.062}{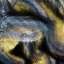}&\myFigure{0.062}{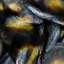}&\myFigure{0.062}{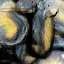}&\myFigure{0.062}{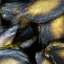}\\
\myFigure{0.062}{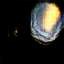}&\myFigure{0.062}{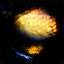}&\myFigure{0.062}{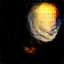}&\myFigure{0.062}{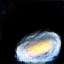}&
\myFigure{0.062}{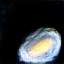}\\
\myFigure{0.062}{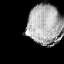}&\myFigure{0.062}{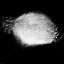}&\myFigure{0.062}{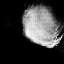}&\myFigure{0.062}{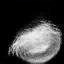}&
\myFigure{0.062}{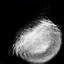}\\\end{tabular}}\\
\end{tabular}
}
\caption{Instance appearances and masks learning results on three selected images of DSF-HOC. From top to bottom, the image rows are real images (only a small portion is showed), the generated composite images by our model, the learned instance appearances by our model and the learned soft mask by our model. See Appendix for more detailed results.}
\label{fig:HOCresults}
\end{figure*}
\subsection{Learning Instance Appearances and Masks From One Image of HOC}
We evaluate our method for appearances and masks learning from one image of HOC on three selected images of DSF-HOC. 
They are an image of the crowd, an image of fruit cluster and an image of seed cluster. They are showed in the first row of Figure  \ref{fig:HOCresults}.
\subsubsection{Data Preparation.}To collect as many patches as possible and reduce the number of instance in each patch, we firstly cut out image patches with a small enough window size. We find that a window size of roughly 1.5 time of the width of the biggest instance works nicely. All patches are scaled into $64\times64$. Both horizontal flip and vertical flip are used for the image of fruit cluster and the image of seed cluster so we can get more patches by taking advantage of isotropy. Patches collected in this way are used as real images for training. The background images are constructed by placing a black disk with radius of roughly 20 pixels in the center of real images, which work better than pure black images. 
\subsubsection{Layout Details.}To layout generated instances in a structure-free way, we firstly pad generated images and then randomly crop them into $64\times64$. The padding sizes for the crowd image, fruit cluster image and seed cluster images are set to 30 pixels, 32 pixels and 26 pixels.
\subsubsection{Results.}
The results are shown in Figure \ref{fig:HOCresults}. Even if there are densely distributed instances in the original images, the generators output only one single instance for each RGBA image. The object surface of every inferred instance is completed, unobscured and realistic even though the most of instances in the original image are more or less occluded. And the generated object-like masks show highly agreement to the corresponding appearance. We believe this demonstrates great potential to
extend our method to unsupervisedly learning of instance generation, segmentation, detection and counting with only a single image of DSF-HOC.
In addition, we find setting padding size to 32 pixels (the fruit cluster case) leads to best alignment and centering for instances in generated images. and there are much more position variance when padding size is 26 pixels (the seed case). Intuitively, the generators may try to compensate for the insufficient poses provided by layout module. 
\section{Conclusion and Future Works}
We proposed a graphical model for natural images capturing dependencies of object appearances, categories and poses. And we pointed out that the conditional independence of object appearances is an important prior for image segmentation. Moreover, a framework was proposed for learning instance segmentation masks with low cost. Our first application provides a new low-cost solution for category-specific foreground segmentation. The generated soft masks are more suitable for image composition than the masks from discriminative methods because of our direct optimization of the composite quality. More impressively, our second application learns object appearances and soft masks with only one image of DSF-HOC without any annotation. It shows a possibility of extremely low-cost unsupervised learning of segmentation, detection, counting, etc. by transferring the knowledge from the instance generators. Our framework also provides a new perspective on universal perceptual grouping that  it can be treated as separating the inputs into components with independent factors. In the future, we plan to develop a trainable multiple objects layout module by advancing the ST-GAN \cite{lin2018st} and design an effective way to jointly train the instance providers and layout module end-to-end.
\section{Acknowledgments}
We thank Yuhao Lu for the help on computational resource in the early time of this work.
\bibliographystyle{aaai}
\bibliography{my_draft.v2}

\begin{thebibliography}{}

\bibitem[\protect\citeauthoryear{Abadi \bgroup et al\mbox.\egroup
  }{2016}]{Abadi2016TensorFlow}
Abadi, M.; Barham, P.; Chen, J.; Chen, Z.; Davis, A.; Dean, J.; Devin, M.;
  Ghemawat, S.; Irving, G.; Isard, M.; et~al.
\newblock 2016.
\newblock Tensorflow: a system for large-scale machine learning.
\newblock In {\em OSDI}, volume~16,  265--283.

\bibitem[\protect\citeauthoryear{Ahn and Kwak}{2018}]{ahn2018learning}
Ahn, J., and Kwak, S.
\newblock 2018.
\newblock Learning pixel-level semantic affinity with image-level supervision
  for weakly supervised semantic segmentation.
\newblock In {\em CVPR}.

\bibitem[\protect\citeauthoryear{Belghazi \bgroup et al\mbox.\egroup
  }{2018}]{Belghazi2018MINE}
Belghazi, I.; Rajeswar, S.; Baratin, A.; Hjelm, R.~D.; and Courville, A.
\newblock 2018.
\newblock Mine: mutual information neural estimation.
\newblock In {\em ICLR}.

\bibitem[\protect\citeauthoryear{Brakel and Bengio}{2017}]{brakel2017learning}
Brakel, P., and Bengio, Y.
\newblock 2017.
\newblock Learning independent features with adversarial nets for non-linear
  ica.
\newblock {\em arXiv preprint arXiv:1710.05050}.

\bibitem[\protect\citeauthoryear{Chen \bgroup et al\mbox.\egroup
  }{2014}]{Chen2014Semantic}
Chen, L.~C.; Papandreou, G.; Kokkinos, I.; Murphy, K.; and Yuille, A.~L.
\newblock 2014.
\newblock Semantic image segmentation with deep convolutional nets and fully
  connected crfs.
\newblock {\em Computer Science} (4):357--361.

\bibitem[\protect\citeauthoryear{Dalal and Triggs}{2005}]{Dalal2005Histograms}
Dalal, N., and Triggs, B.
\newblock 2005.
\newblock Histograms of oriented gradients for human detection.
\newblock In {\em CVPR}, volume~1,  886--893.

\bibitem[\protect\citeauthoryear{Eslami \bgroup et al\mbox.\egroup
  }{2016}]{Eslami2016Attend}
Eslami, S.~A.; Heess, N.; Weber, T.; Tassa, Y.; Szepesvari, D.; Hinton, G.~E.;
  et~al.
\newblock 2016.
\newblock Attend, infer, repeat: Fast scene understanding with generative
  models.
\newblock In {\em NIPS},  3225--3233.

\bibitem[\protect\citeauthoryear{Goodfellow \bgroup et al\mbox.\egroup
  }{2014}]{Goodfellow2014Generative}
Goodfellow, I.; Pouget-Abadie, J.; Mirza, M.; Xu, B.; Warde-Farley, D.; Ozair,
  S.; Courville, A.; and Bengio, Y.
\newblock 2014.
\newblock Generative adversarial nets.
\newblock In {\em NIPS},  2672--2680.

\bibitem[\protect\citeauthoryear{Greff \bgroup et al\mbox.\egroup
  }{2016}]{Greff2016Tagger}
Greff, K.; Rasmus, A.; Berglund, M.; Hao, T.; Valpola, H.; and Schmidhuber, J.
\newblock 2016.
\newblock Tagger: Deep unsupervised perceptual grouping.
\newblock In {\em NIPS},  4484--4492.

\bibitem[\protect\citeauthoryear{G{\"u}{\c{c}}l{\"u} \bgroup et al\mbox.\egroup
  }{2017}]{BU2LearnCRF}
G{\"u}{\c{c}}l{\"u}, U.; G{\"u}{\c{c}}l{\"u}t{\"u}rk, Y.; Madadi, M.; Escalera,
  S.; Bar{\'o}, X.; Gonz{\'a}lez, J.; van Lier, R.; and van Gerven, M.~A.
\newblock 2017.
\newblock End-to-end semantic face segmentation with conditional random fields
  as convolutional, recurrent and adversarial networks.
\newblock {\em arXiv preprint arXiv:1703.03305}.

\bibitem[\protect\citeauthoryear{Hu \bgroup et al\mbox.\egroup
  }{2018}]{hu2018relation}
Hu, H.; Gu, J.; Zhang, Z.; Dai, J.; and Wei, Y.
\newblock 2018.
\newblock Relation networks for object detection.
\newblock In {\em CVPR}, volume~2.

\bibitem[\protect\citeauthoryear{Huang and Murphy}{2015}]{huang2015efficient}
Huang, J., and Murphy, K.
\newblock 2015.
\newblock Efficient inference in occlusion-aware generative models of images.
\newblock {\em arXiv preprint arXiv:1511.06362}.

\bibitem[\protect\citeauthoryear{Johnson, Gupta, and
  Fei-Fei}{2018}]{johnson2018image}
Johnson, J.; Gupta, A.; and Fei-Fei, L.
\newblock 2018.
\newblock Image generation from scene graphs.
\newblock In {\em CVPR}.

\bibitem[\protect\citeauthoryear{Khoreva \bgroup et al\mbox.\egroup
  }{2017}]{BU3GrabCut}
Khoreva, A.; Benenson, R.; Hosang, J.~H.; Hein, M.; and Schiele, B.
\newblock 2017.
\newblock Simple does it: Weakly supervised instance and semantic segmentation.
\newblock In {\em CVPR}, volume~1, ~3.

\bibitem[\protect\citeauthoryear{Kingma and Ba}{2014}]{Kingma2014Adam}
Kingma, D.~P., and Ba, J.
\newblock 2014.
\newblock Adam: A method for stochastic optimization.
\newblock {\em arXiv preprint arXiv:1412.6980}.

\bibitem[\protect\citeauthoryear{Kingma and Welling}{2013}]{Kingma2013Auto}
Kingma, D.~P., and Welling, M.
\newblock 2013.
\newblock Auto-encoding variational bayes.
\newblock In {\em ICLR}.

\bibitem[\protect\citeauthoryear{Lin \bgroup et al\mbox.\egroup
  }{2018}]{lin2018st}
Lin, C.-H.; Yumer, E.; Wang, O.; Shechtman, E.; and Lucey, S.
\newblock 2018.
\newblock St-gan: Spatial transformer generative adversarial networks for image
  compositing.
\newblock In {\em CVPR},  9455--9464.

\bibitem[\protect\citeauthoryear{Liu \bgroup et al\mbox.\egroup
  }{2015}]{liu2015faceattributes}
Liu, Z.; Luo, P.; Wang, X.; and Tang, X.
\newblock 2015.
\newblock Deep learning face attributes in the wild.
\newblock In {\em ICCV}.

\bibitem[\protect\citeauthoryear{Long, Shelhamer, and
  Darrell}{2015}]{long2015fully}
Long, J.; Shelhamer, E.; and Darrell, T.
\newblock 2015.
\newblock Fully convolutional networks for semantic segmentation.
\newblock In {\em CVPR},  3431--3440.

\bibitem[\protect\citeauthoryear{Miyato \bgroup et al\mbox.\egroup
  }{2018}]{Miyato2018Spectral}
Miyato, T.; Kataoka, T.; Koyama, M.; and Yoshida, Y.
\newblock 2018.
\newblock Spectral normalization for generative adversarial networks.
\newblock In {\em ICLR}.

\bibitem[\protect\citeauthoryear{Newell and Deng}{2017}]{newell2017pixels}
Newell, A., and Deng, J.
\newblock 2017.
\newblock Pixels to graphs by associative embedding.
\newblock In {\em NIPS},  2171--2180.

\bibitem[\protect\citeauthoryear{Nowozin, Cseke, and
  Tomioka}{2016}]{Nowozin2016f}
Nowozin, S.; Cseke, B.; and Tomioka, R.
\newblock 2016.
\newblock f-gan: Training generative neural samplers using variational
  divergence minimization.
\newblock In {\em NIPS},  271--279.

\bibitem[\protect\citeauthoryear{Pathak \bgroup et al\mbox.\egroup
  }{2017}]{Pathak2017Learning}
Pathak, D.; Girshick, R.~B.; Doll{\'a}r, P.; Darrell, T.; and Hariharan, B.
\newblock 2017.
\newblock Learning features by watching objects move.
\newblock In {\em CVPR}, volume~1, ~7.

\bibitem[\protect\citeauthoryear{Radosavovic \bgroup et al\mbox.\egroup
  }{2018}]{distillation}
Radosavovic, I.; Dollár, P.; Girshick, R.; Gkioxari, G.; and He, K.
\newblock 2018.
\newblock Data distillation: Towards omni-supervised learning.
\newblock In {\em CVPR}.

\bibitem[\protect\citeauthoryear{Ronneberger, Fischer, and
  Brox}{2015}]{Ronneberger2015U}
Ronneberger, O.; Fischer, P.; and Brox, T.
\newblock 2015.
\newblock U-net: Convolutional networks for biomedical image segmentation.
\newblock In {\em International Conference on Medical image computing and
  computer-assisted intervention},  234--241.

\bibitem[\protect\citeauthoryear{Roy and Todorovic}{2017}]{BU1}
Roy, A., and Todorovic, S.
\newblock 2017.
\newblock Combining bottom-up, top-down, and smoothness cues for weakly
  supervised image segmentation.
\newblock In {\em CVPR},  7282--7291.

\bibitem[\protect\citeauthoryear{Selvaraju \bgroup et al\mbox.\egroup
  }{2017}]{selvaraju2017grad}
Selvaraju, R.~R.; Cogswell, M.; Das, A.; Vedantam, R.; Parikh, D.; and Batra,
  D.
\newblock 2017.
\newblock Grad-cam: Visual explanations from deep networks via gradient-based
  localization.
\newblock In {\em ICCV},  618--626.

\bibitem[\protect\citeauthoryear{Vondrick, Pirsiavash, and
  Torralba}{2016}]{Vondrick2016Generating}
Vondrick, C.; Pirsiavash, H.; and Torralba, A.
\newblock 2016.
\newblock Generating videos with scene dynamics.
\newblock In {\em NIPS},  613--621.

\bibitem[\protect\citeauthoryear{Wah \bgroup et al\mbox.\egroup
  }{2011}]{wah2011caltech}
Wah, C.; Branson, S.; Welinder, P.; Perona, P.; and Belongie, S.
\newblock 2011.
\newblock The caltech-ucsd birds-200-2011 dataset.
\newblock Technical Report CNS-TR-2011-001, California Institute of Technology.

\bibitem[\protect\citeauthoryear{Wei \bgroup et al\mbox.\egroup
  }{2017}]{Wei2017Object}
Wei, Y.; Feng, J.; Liang, X.; Cheng, M.-M.; Zhao, Y.; and Yan, S.
\newblock 2017.
\newblock Object region mining with adversarial erasing: A simple
  classification to semantic segmentation approach.
\newblock In {\em CVPR}, volume~1, ~3.

\bibitem[\protect\citeauthoryear{Wu \bgroup et al\mbox.\egroup }{2018}]{HOC}
Wu, Z.; Chang, R.; Ma, J.; Lu, C.; and Tang, C.~K.
\newblock 2018.
\newblock Annotation-free and one-shot learning for instance segmentation of
  homogeneous object clusters.
\newblock In {\em IJCAI},  1036--1042.

\bibitem[\protect\citeauthoryear{Yang \bgroup et al\mbox.\egroup
  }{2017}]{Yang2017LR}
Yang, J.; Kannan, A.; Batra, D.; and Parikh, D.
\newblock 2017.
\newblock Lr-gan: Layered recursive generative adversarial networks for image
  generation.
\newblock In {\em ICLR}.

\bibitem[\protect\citeauthoryear{Zhou \bgroup et al\mbox.\egroup
  }{2016}]{Zhou2016Learning}
Zhou, B.; Khosla, A.; Lapedriza, A.; Oliva, A.; and Torralba, A.
\newblock 2016.
\newblock Learning deep features for discriminative localization.
\newblock In {\em CVPR},  2921--2929.

\bibitem[\protect\citeauthoryear{Zhou \bgroup et al\mbox.\egroup
  }{2018}]{BU4MCG}
Zhou, Y.; Zhu, Y.; Ye, Q.; Qiu, Q.; and Jiao, J.
\newblock 2018.
\newblock Weakly supervised instance segmentation using class peak response.
\newblock In {\em CVPR}.

\bibitem[\protect\citeauthoryear{Zhu, Mumford, and
  others}{2007}]{zhu2007stochastic}
Zhu, S.-C.; Mumford, D.; et~al.
\newblock 2007.
\newblock A stochastic grammar of images.
\newblock {\em Foundations and Trends{\textregistered} in Computer Graphics and
  Vision} 2(4):259--362.

\end{thebibliography}
\section{Appendix}
\subsection{Data Distillation in the First Application}
We train a student network to learn the teacher network obtained by adversarial training. The training samples are constructed based on the raw results predicted by the teacher network. In more detail, we feed the student network with the augmented input images that are composited from carved foregrounds (by raw results) and random backgrounds, and the label for training is directly the prediction of teacher network. Additional horizontal flip augmentation is used. We adopt smooth $L_1$ loss for each pixel, which is is less sensitive to outliers and thus helpful to remove the noise caused by unstable adversarial training.
\begin{equation}
smooth_{L1}(x)=
\begin{cases}
4x^2,& \mbox{if $|x|<0.25$}\\
|x|,& \mbox{otherwise}
\end{cases}
\end{equation}
\subsection{Additional Experiments Details}
In the second application, we use an information regularization term like InfoGAN for more diversity of generation results. The weight for this term is denoted as $\beta$.
\begin{table}[!htbp]
  \tiny
  \centering
  \caption{The networks in the first application}
  \label{table0}
  \centering
  \begin{tabular}{lll}
    \cmidrule(r){1-2}
    Discriminator     & UNet \\
    \midrule
    Input 128x128x3 image & Input 128x128x3 \\
    3x3 conv. 64 lRELU. stride 2. SN  & 4x4 conv. 64 lRELU. stride 2. BN \\
    3x3 conv. 128 lRELU. stride 2. SN & 4x4 conv. 128 lRELU. stride 2. BN \\
    3x3 conv. 256 lRELU. stride 2. SN & 4x4 conv. 256 lRELU. stride 2. BN \\
    3x3 conv. 512 lRELU. stride 2. SN & 4x4 conv. 512 lRELU. stride 2. BN \\
    3x3 conv. 256 lRELU. stride 1. SN & 4x4 conv. 512 lRELU. stride 2. BN \\
    1x1 conv. 1 lRELU+sigmoid. stride 1. SN & 2x2 conv. 512 lRELU. stride 2. BN\\
    & 2x2 deconv. 512 lRELU. stride 2. BN\\
    & 4x4 deconv. 512 lRELU. stride 2. BN\\
    & 4x4 deconv. 256 lRELU. stride 2. BN\\
    & 4x4 deconv. 128 lRELU. stride 2. BN\\
    & 4x4 deconv. 64 lRELU. stride 2. BN\\
    & 4x4 deconv. 1 sigmoid. stride 2. \\
    \bottomrule
  \end{tabular}
\end{table}
\begin{table}[!htbp]
  \footnotesize
  \centering
  \caption{The networks in the second application}
  \label{sample-table}
  \tiny
  \begin{tabular}{lll}
    \cmidrule(r){1-2}
    Discriminator     & Genearator \\
    \midrule
    Input 64x64x3 image & Input 1x1x6 z \\
    3x3 conv. 64 lRELU. stride 2. SN  & 1x1 conv. 8*8*256 lRELU. stride 1. BN \\
    3x3 conv. 128 lRELU. stride 2. SN & 4x4 conv. 256 lRELU. stride 1. BN \\
    3x3 conv. 256 lRELU. stride 2. SN & 4x4 conv. 256 lRELU. stride 1. BN \\
    3x3 conv. 512 lRELU. stride 2. SN & 4x4 deconv. 128 lRELU. stride 2. BN \\
    3x3 conv. 1024 lRELU. stride 1. SN & 4x4 deconv. 64 lRELU. stride 2. BN \\
    1x1 conv. 1 lRELU+sigmoid. stride 1. SN & 4x4 deconv. 4 sigmoid. stride 2.\\
    \bottomrule
  \end{tabular}
\end{table}
\begin{table}[!htbp]
  \centering
  \caption{The hyperparameters in the first application}
  \label{table 2}
  \centering
  \begin{tabular}{lll}
    \cmidrule(r){1-3}
    dataset & $\lambda$ & $a$ \\
    \midrule
    CelebA & 1000 & 0.25\\
    Caltech-200 bird & 1000 & 0.25\\
    \bottomrule
  \end{tabular}
\end{table}
\begin{table}[!htbp]
 \tiny
  \centering
  \caption{The hyperparameters in the second application}
  \label{table 2}
  \centering
  \begin{tabular}{llllll}
    \cmidrule(r){1-6}
    Dataset & \# Composition layers & $\lambda$ & dim$(\bm{z})$ & $a$ & $\beta$\\
    \midrule
    crowd & 8 & 1000 & 6 & 0.1 & 50\\
    fruit cluster & 10 & 1000 & 5 & 0.25 & 50\\
    seed cluster & 12 & 1000 & 2 & 0.25 & 50\\
    \bottomrule
  \end{tabular}
\end{table}
\newpage
\subsection{Additional Results}
\begin{figure}[!hp]
  \centering
  \includegraphics[width=0.5\linewidth]{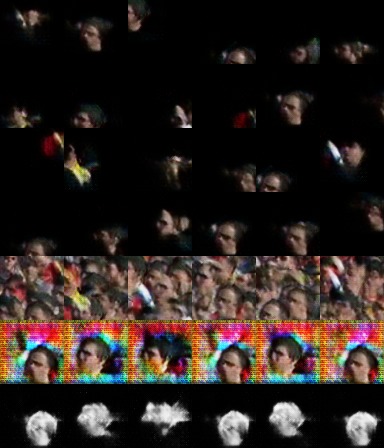}\includegraphics[width=0.5\linewidth]{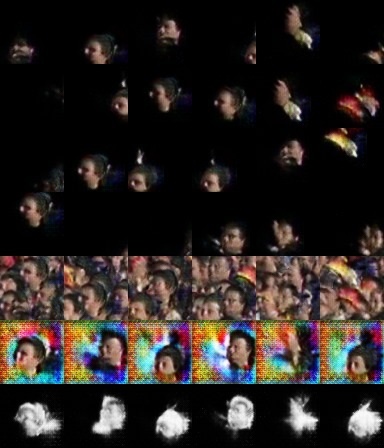}
 \label{appendix_crowd}
 \caption{Additional results of appearances and masks learning on one image of the crowd. The first four rows are carved instances for final composition (only the last four image layers are showed) , the fifth row is generated composite image based on the image layers as shown in previous rows, the sixth row is the learned
instance appearances and the last row is the learned soft masks by our model.}
\end{figure}

\begin{figure}[!hp]
  \centering
  \includegraphics[width=0.5\linewidth]{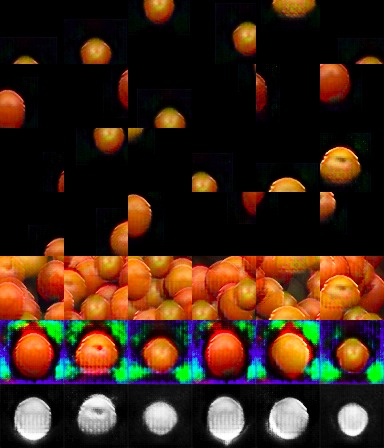}\includegraphics[width=0.5\linewidth]{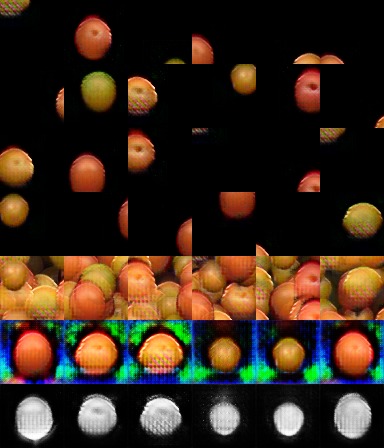}
 \label{appendix_crowd}
 \caption{Additional results of appearances and masks learning on one image of fruit cluster. The first four rows are carved instances for final composition (only the last four image layers are showed) , the fifth row is generated composite image based on the image layers as shown in previous rows, the sixth row is the learned
instance appearances and the last row is the learned soft masks by our model.}
\end{figure}
 
\begin{figure*}[!hp]
  \centering
\includegraphics[width=0.33\linewidth]{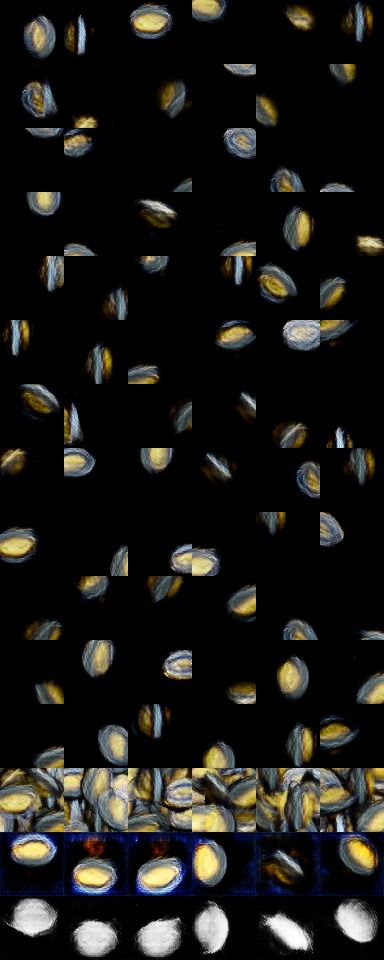}\includegraphics[width=0.33\linewidth]{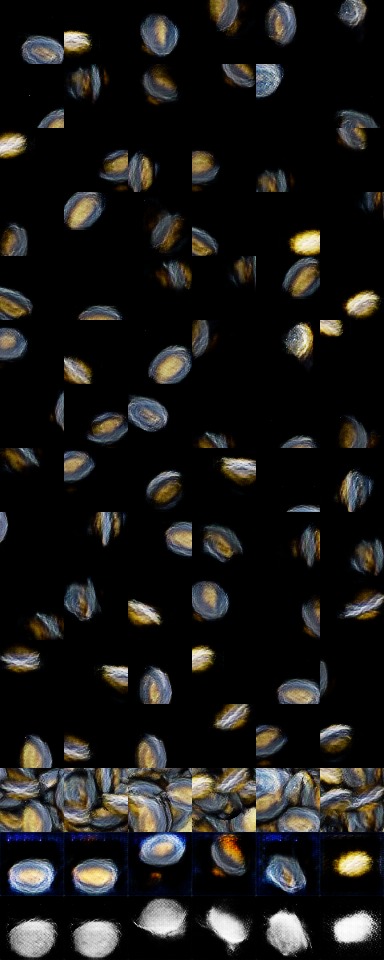}\includegraphics[width=0.33\linewidth]{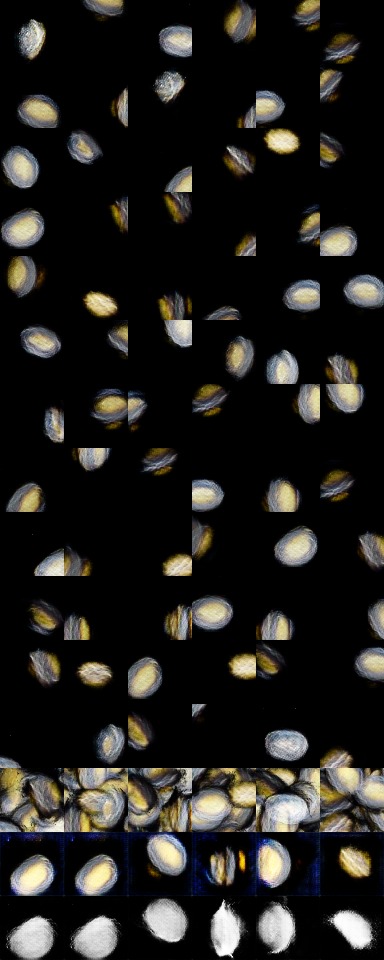}
 \label{appendix_seed}
 \caption{Additional results of appearances and masks learning on one image of seed cluster. The first twelve rows are carved instances for final composition (all image layers are showed), the third row from the bottom is generated composite image based on the previous rows, the second row from the bottom row is the learned
instance appearances and the last row is the learned soft masks by our model.}
\end{figure*}

\begin{figure*}[!htbp]
  \centering
  \includegraphics[width=\linewidth]{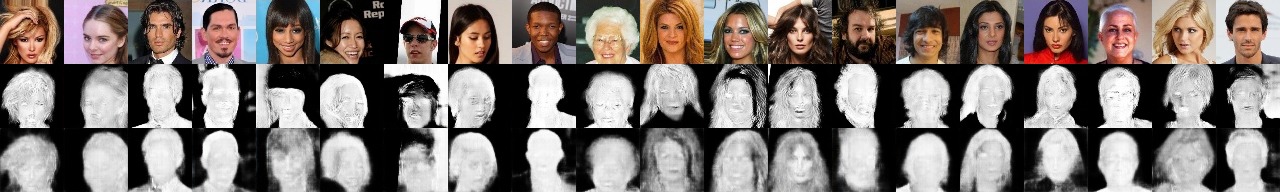}\\
  \includegraphics[width=\linewidth]{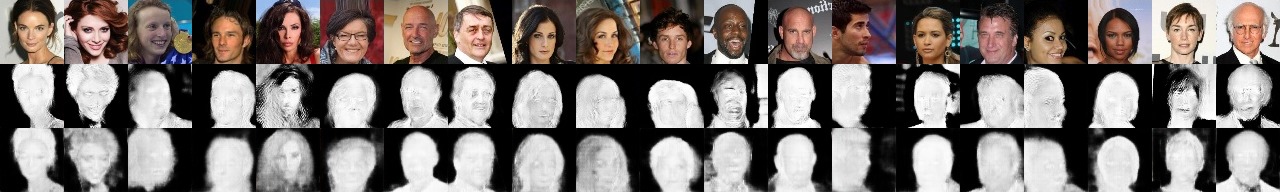}\\
  \includegraphics[width=\linewidth]{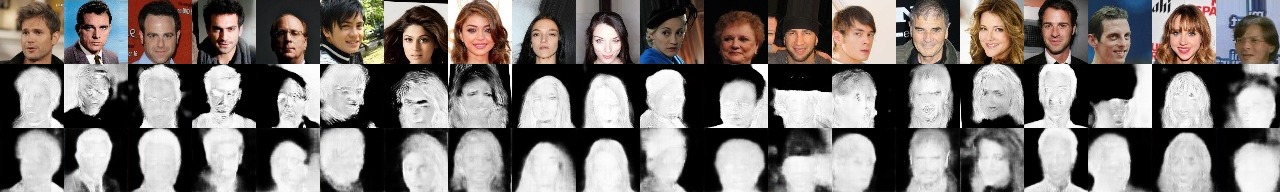}\\
  \includegraphics[width=\linewidth]{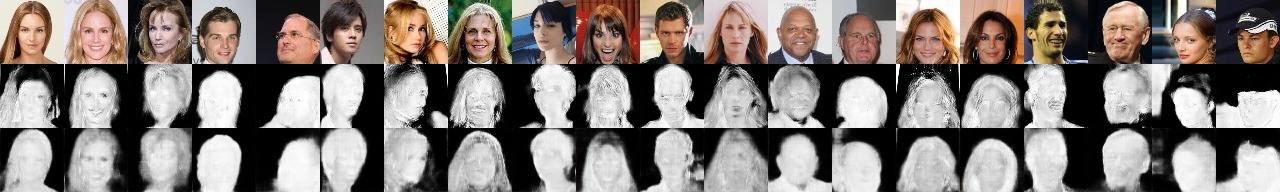}\\
   \includegraphics[width=\linewidth]{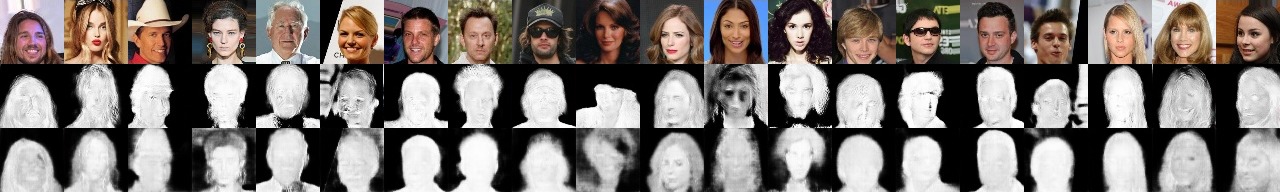}\\
 \label{appendix_celeba}
 \caption{Additional foreground segmentation results on CelebA dataset. The first rows are input images, the second rows are the raw results obtained by adversarial training and the third rows are the results with data distillation.}
\end{figure*}

\begin{figure*}[!htbp]
  \centering
  \includegraphics[width=\linewidth]{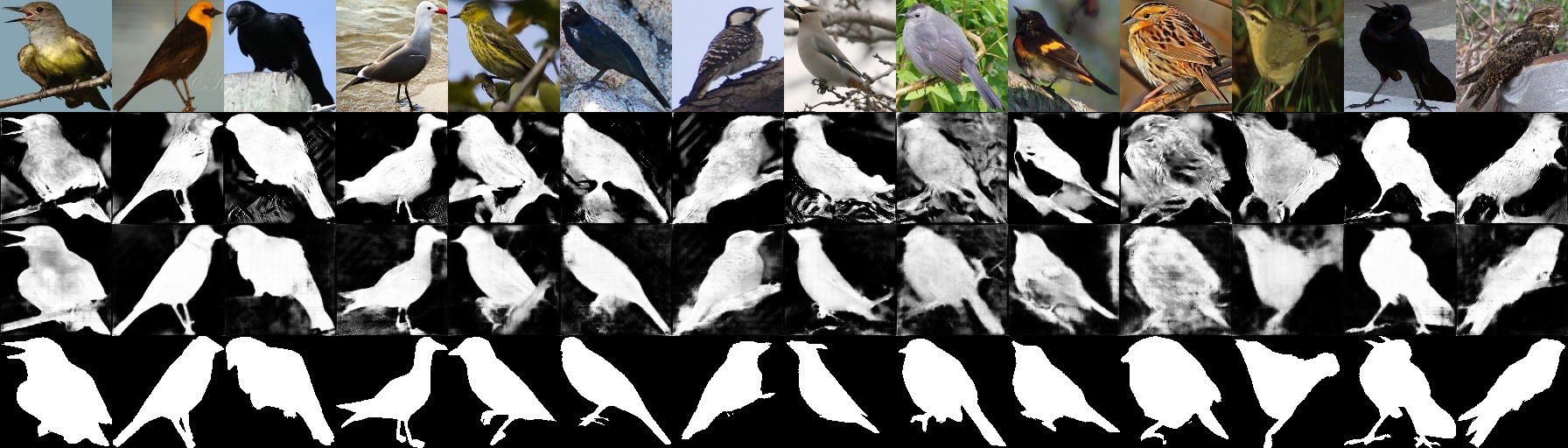}\\
  \includegraphics[width=\linewidth]{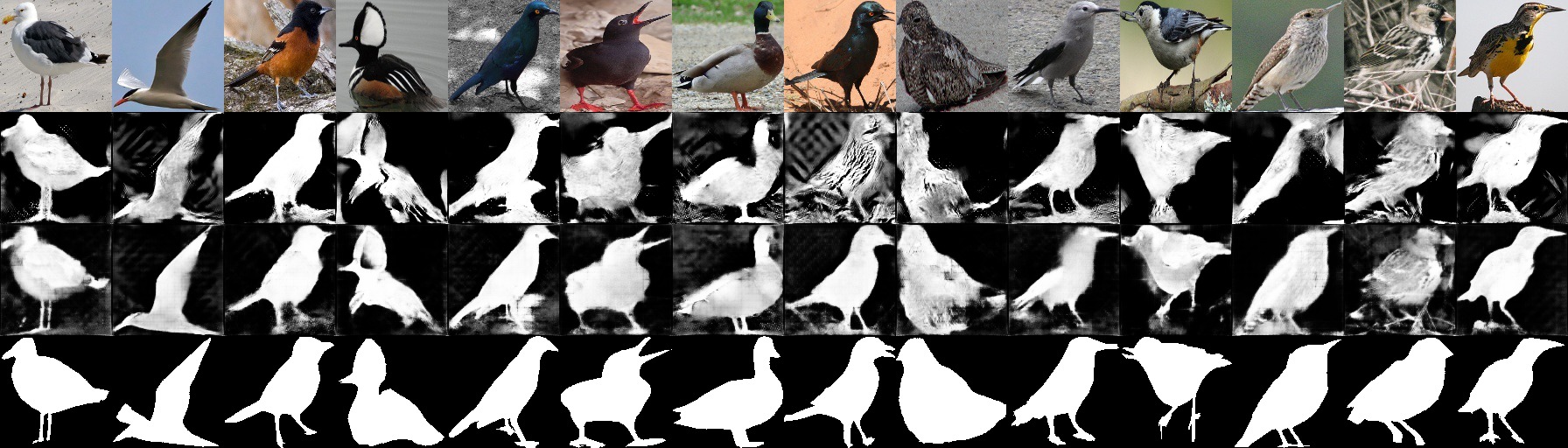}\\
 \caption{Additional foreground segmentation results on Caltech-200 bird dataset. The first rows are input images, the second rows are the raw results obtained by adversarial training, the third rows are the results with data distillation and the fourth rows are the ground truth.}
\end{figure*}
\end{document}